# Bio-inspired Optimization: metaheuristic algorithms for optimization

Pravin S. Game
Research Scholar, Shri. JJT University,
Jhunjhunu, Rajasthan, India
pravinsgame@gmail.com

Dr. Vinod Vaze
Shri. JJT University
Jhunjhunu, Rajasthan, India

Dr. Emmanuel M.
Department of Information Technology, Pune Institute of Computer Technology, Pune.

*Abstract*— In today's day and time solving real-world complex problems has become fundamentally vital and critical task. Many of these are combinatorial problems, where optimal solutions are sought rather than exact solutions. Traditional optimization methods are found to be effective for small scale problems. However, for real-world large scale problems, traditional methods either do not scale up or fail to obtain optimal solutions or they end-up giving solutions after a long running time. Even earlier artificial intelligence (AI) based techniques used to solve these problems could not give acceptable results. However, last two decades have seen many new methods in AI based on the characteristics and behaviors of the living organisms in the nature which are categorized as bio-inspired or nature inspired optimization algorithms. These methods, are also termed metaheuristic optimization methods, have been proved theoretically and implemented using simulation as well used to create many useful applications. They have been used extensively to solve many industrial and engineering complex problems due to being easy to understand, flexible, simple to adapt to the problem at hand and most importantly their ability to come out of local optima traps. This local optima avoidance property helps in finding global optimal solutions.

This paper is aimed at understanding how nature has inspired many optimization algorithms, basic categorization of them, major bio-inspired optimization algorithms invented in recent time with their applications.

*Keywords*— *optimization methods, bio-inspired, applications, evolutionary, metaheuristic.*

I. INTRODUCTION

Historically, it has been observed that human beings are always intrigued by the nature and its functioning. Nature is very complex. From seeing the nature only as source of food and shelter, humans started seeing it objectively as well as subjectively. Various physical phenomena, food foraging behaviors of insects and animals, socio-cultural conducts of insects and animals were observed and studied by the humans to better understand the functioning of nature, to manage nature, to adapt to natural conditions, then to create genetically modified foods, seeds, and animals, and even to control spread of diseases using vaccines.

However, things started changing when the computing machines (computers) were invented and were evolved to become better and better in computational power and efficiency. Computer was used to solve the complex optimization problems using traditional methods, however, had its limitations. Now, nature was seen as de-facto source of inspiration for building new algorithms applicable to solve such complex optimization problems. Such systems build using nature inspired algorithms were called nature-inspired systems or bio-inspired systems or natural computing systems. Many even termed it as a new revolution in computing – natural computing revolution. This bio-inspired computing has started reducing the gap between computing and nature. This natural computing has been divided into three major classes by [1], viz. a) computing inspired by nature b) simulation and emulation of nature c) computing with natural materials. In this natural computing various nature based algorithms are developed and used. As said these nature based algorithms are metaheuristic algorithms which are also classified into three parts [2], viz. a) evolutionary algorithms, b) Physics-based algorithms and c) swarm intelligence algorithms. It should be noted that these three types of algorithms fall in the first category of natural computing i.e. computing inspired by nature.

The rest of the paper is organized as follows: sections II briefs the three natural computing classes, section III presents the review of various categorized recent algorithms, and section IV presents the conclusions.

II. NATURAL COMPUTING

Over the period of time, it is understood that though nature is complex, its processes and functions can be modeled as some set of basic finite rules and parameters which can then be transferred for computing. The entire domain of bioinspired computing is based on searching for these basic rules governing the natural system under consideration. Once these set of rules are identified, these are then mapped to the corresponding computing paradigm. Let us briefly understand this through following sections.

*A. Computing inspired by nature*

This is the oldest and most researched approach of natural computing with two major objectives: 1) build theoretical model of natural phenomenon and simulate using computers, and 2) develop alternative algorithms to solve complex problems which otherwise could not be solved satisfactorily by traditional methods.

The earliest pioneering work in this direction could be attributed to the development of a logical model of neuron by [3] in 1943. Here, propositional logic was used to represent the neurons, neural events and their relationships. This model of neural nets, in later times, gave rise to a field of study called artificial neural networks (ANN).

In the following decades many works based on Darwin's theory of evolution which talks of survival of the fittest were developed. Three principles of this evolution theory are population reproduces, genetic varies and selection of fittest. Such biology inspired computing approaches are called as evolutionary computing where evolutionary algorithms (EAs) are developed to solve search and optimization problems. Then there is other class in this where physics-based phenomenon are studied and modeled. For example simulated annealing [4], inspired from the statistical thermodynamics for annealing of solids, was claimed to be very useful in solving complex combinatorial optimization problems. Then based the behaviors of birds and animals many swarm based algorithms were developed creating a new category of nature inspired computing called as swarm intelligence (SI) algorithms. More on EAs, physics-based and SI algorithms will be discussed in section III.

*B. Simulation and Emulation of nature in computer*

Nature is full of fractals, for example, mountains, tree leaves, coastlines, cauliflower, brain, lungs, kidney are all fractals. In computer graphics, fractals can be emulated using fractal geometry. Such models when simulated or emulated by computers become very useful not only in analyzing a natural phenomenon but also can be extended for cancer cell identification, binding of antibodies to antigens, detailing forest fires etc. But more interesting field of these simulations & emulations is found in creation of artificial life. Artificial life deals with creation of systems mimicking the behaviors of organisms or phenomenon in order to synthesize it and understand it. The major goal here is to create living systems from non-living parts. An example of this is Artificial Intelligence robot (AIBO) [5], emulated robotic pet dogs. Another example where natural concept is implemented for computer is virus. Computer virus is metaphor for virus from nature, having similar properties.

*C. Computing with natural material*

With the limitation of spacing transistors in integrated circuits, the obvious question would be can silicon be replaced by other material? Nature comes to rescue from this question by providing alternate solutions for computing. There is lot of research going on for identifying the solutions based on nature specifically DNA computing and quantum computing. For DNA computing was used to find square roots of numbers [6] and quantum computing based service was recently launched by Amazon web service [7].

III. BIO-INSPIRED METAHEURISTIC OPTIMIZATION APPROACHES

The relevant definition of heuristic from Merriam-Webster dictionary is 'of or relating to exploratory problem-solving techniques that utilize self-educating techniques to improve performance'. These algorithms use some available information for taking next step towards the solution. Metaheuristic is a heuristic defined to select a heuristic providing better solutions. Metaheuristics are generally problem independent, and can be of following types:

*a)* local search based and global search based.

*b)* single-solution based and population-based

*c)* with memory and memoryless

*d)* greedy and iterative

*e)* Parallel

*f)* Nature-inspired and hybridized

Fig.1 is a good representative of the metaheuristics classification [8].

Metaheuristic optimization methods with inspiration from nature have led to discovery of large number of optimization algorithms in various categories as detailed below.

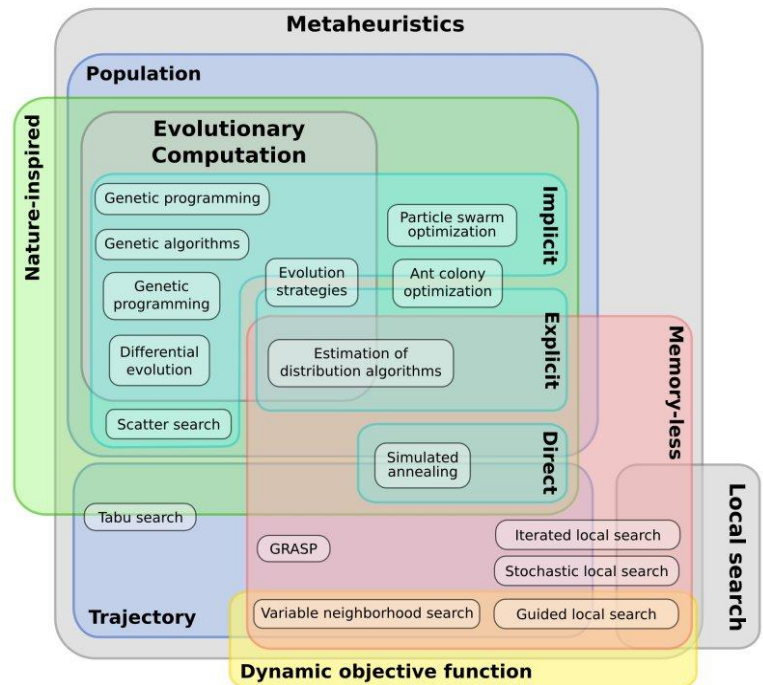

Fig. 1 Euler's diagram of the different classifications of metaheuristics

*A. Evolutionary algorithms*

Darwin's theory of evolution, though radical and controversial, initiated a lot of research and led to development of lot of nature inspired algorithms for evolutionary computing. EAs consider that finding a solution for the problem is nothing but a search task, in which starting with a random solution from solution space; algorithm is employed to search for better solutions than the already considered one. Genetic algorithms (GA) are foremost in this category where chromosomes were used to code and process the information [9, 10, 11, 12]. These chromosomes, also called candidate solutions, are processed by various genetic operators, viz. reproduction, crossover, mutation, and selection until termination criteria is reached. Termination criteria could be maximum number of iterations, convergence of solution or reaching a best solution.

Evolutionary strategies (ES) are credited to German researchers Rechenberg and Schwefel [13,14], considered as black box optimization technique are based on natural evolution. ES also has a population which is iterated as generation. In each generation, the individual, every genotype is evaluated based on a fitness function, best are kept and remaining are discarded. Those retained genotypes are mutated for improving the fitness. The fittest are selected as population of next iteration (generation) and the process is repeated to get optimized results. ES have proved to be good algorithms to solve many optimization problems. This class of EAs has continued to attract researchers [15, 16, 17, 18, 19, 20].

As an alternative to AI, evolutionary programming (EP) was invented by Fogel and others in 1966 [21]. Finite state machine (FSM) was employed in this. Finite alphabet produces finite sequence of symbols which creates an environment. Intelligent behavior considers prediction of this environment. With a well-defined evaluation function, algorithm is evolved by processing this set of sequences to produce new output from the current state and current input. As it is well known that the FSM provides a good predictive mechanism for set of inputs, its use for evolving the algorithm to improve the performance of algorithm is appropriate. In evolutionary programming selection operator is based on tournament selection method and for mutation operator is Gaussian mutation. The procedure of EP is similar to EA. It has been quite useful in pattern recognition and classification.

David Goldberg, a student of John Holland, coined the term genetic programming (GP). Two seminal works on genetic programming were published by the students of Holland in 1985 by Cramer [22] and in 1988 by Koza [23]. It's class of GA where programs are the initial population. GP works on these set of programs and evolution of programs is done by using selection, crossover, and mutation operators. Programs are evaluated for their fitness in each iteration and evolved to get optimal program over the iterations. This EA category has also been widely researched and various types of GP are developed, like tree-based GP, stack-based GP, Cartesian GP etc. [24].

Differential Evolution proposed in [25] is a direct search method, inherently parallel, finds true global minimum with faster convergence. It starts with creation of random population and generation of trial parameter vectors. Random variation is added to third vector by taking weighted difference between two random parameter vectors. This newly created random vector is then compared with the selected vector and it is better then added to the population in next iteration.

Other two EAs of interest here are Biogeography-Based optimization (BBO) [26] and Population-Based Incremental Learning (PBIL) [27, 28].

### B. Physics-based optimization algorithms

Table 1 gives a list of physics-based algorithms and the related physics law.

TABLE 1 MAPPING OF OPTIMIZATION ALGORITHM WITH LAWS OF PHYSICS

| Optimization algorithm | Law(s)/concepts of Physics |
| --- | --- |
| Hysteresis Optimization [29] | Demagnetization process |
| Gravitational Local Search [30] | Law of Gravitation |
| Electromagnetism-like Algorithm [31] | Electromagnetism law |
| Space Gravitational Algorithm [32] | Theories of relativity and gravity |
| Particle Collision Algorithm [33] | Nuclear collision reactions |
| Small World Optimization Algorithm [34] | Small world phenomenon |
| Big Bang Big Crunch [35] | Big bang theory and big-crunch theory |
| Central Force Optimization [36] | Gravitational kinematics |
| Integrated Radiation Algorithm [37] | Gravitational radiation |
| Big Crunch Algorithm [38] | Closed universe theory |
| Gravitational Search Algorithm [39] | Laws of gravity and motion |
| River formation dynamics algorithm [40] | Ground erosion and sediment deposits |
| Intelligent Water Drops [41] | Water drops moving as a big swarm in river |
| Artificial Physics Algorithm [42] | Physicomimetics framework and physical forces |
| Light Ray Optimization [43] | Optical refraction and reflection of light rays |
| Charged System Search algorithm [44] | Gauss and Coulomb laws |
| Gravitation Field Algorithm [45] | Solar Nebula Disk Model |
| Artificial Chemical Reaction Optimization Algorithm [46] | Types and occurring of chemical reactions |
| Galaxy based Search Algorithm [47] | Spiral arms of spiral galaxies |
| Magnetic Optimization Algorithm [48] | Theory of magnetism |
| Gravitaional Interactions Optimization [49] | Laws of gravity and motion |
| Spiral Optimization Algorithm [50] | Spiral phenomenon |
| Water Flow Algorithm [51] | Hydrological cycle of meteorology and erosion phenomenon |
| Black Hole [52] | Black hole phenomenon |
| Ray Optimization [53] | Snell's law |
| Curve Space Optimization [54] | General relativity theory |
| Water Cycle Aalgorithm [55] | Condensation and evaporation of water |
| Ion Motion Optimization [56] | Push & pull forces of anions and cations |
| Tug of war optimization [57] | Newton's laws of mechanics |

### C. Swarm Intelligence based algorithms

This category of bioinspired metaheuristic has seen rampant development of numerous algorithms. Swarm intelligence (SI) represents the emergent properties from the interactions of the group of simple individuals or agents. It considers that synergetic behaviors of agents provide better solutions than the individual ones. SI has its advantages over evolutionary algorithms in terms of fewer parameters, memorization of past best results, lesser operators and ease of implementation.

Swarm intelligence metaheuristic are inspired mostly from the insect and bird colonies, flocks of birds, school of fishes, herds of animals. Earliest used SI methods include ant colony optimization (ACO) [58], and particle swarm optimization (PSO) [59]. ACO and PSO have been used in solving many optimization problems including travelling salesman problem and are generally used for comparing the results of newly proposed swarm algorithms. ACO is based on the food foraging behaviors of ants and has seen wide applicability like for route optimization, pattern recognition, classification and clustering among others. Ants remember the path travelled by the level of pheromone. PSO assigns a random velocity to each particle (candidate solution) and then these particles are flown through the solution space (hyperspace). PSO keeps tracks of best fitness obtained globally (called gbest) and each particle remembers the best solution (called pbest) it has obtained so far. Algorithms proceeds by accelerating the particles towards the gbest and pbest. PSO is also widely used in application ranging from engineering to medical sciences. Last two decade has seen a rising interest of researchers in developing metaheuristic algorithms taking inspiration from the evolution or various behaviors of insects, birds and animals. Table 2 lists down some prominent algorithms in this category.

TABLE 2 SI ALGORITHMS

| Optimization algorithm | Behavior(s) considered |
|---|---|
| Bat Algorithm [60] | Echo-cancellation |
| Cuckoo search [61] | Obligate brood parasitic behavior |
| Flower-pollination algorithm [62] | Pollination process of flowers |
| Fire-fly algorithm [63] | Flashing light patterns |
| Moth-flame optimization [64] | Spiral flying path of moth |
| Glowworm Swarm Optimization [65] | Luciferin induced glowing behavior |
| Bees Algorithm [66] | Food foraging behavior of honeybees |
| Artificial Bee Colony [67] | Swarming around hive by honey bees |
| Cuckoo optimization algorithm [68] | Cuckoos' survival efforts |
| Grey wolf optimizer [69] | Hunting behavior & social hierarchy |
| Dolphin echolocation [70] | Echolocation ability |
| Ant-lion optimizer [71] | Hunting mechanism |
| Whale optimization algorithm [72] | Bubble-net hunting |
| Fruit fly optimization [73] | Food foraging behavior |
| Hunting search [74] | Group hunting behavior |
| Salp swarm algorithm [75] | Navigation and foraging behaviors |
| Grasshopper optimization algorithm [76] | Social interaction and food foraging |
| Dragonfly algorithm [77] | Static and dynamic swarming behavior |
| Deer hunting optimization [78] | Hunting behavior of humans |
| SailFish optimizer [79] | Hunting behavior |
| Barnacles mating optimizer [80] | Mating behavior |
| Manta ray foraging optimizer [81] | Foraging behaviors- chain, cyclone, somersault |

## IV. DISCUSSION AND CONCLUSIONS

Irrespective of type of metaheuristic, search for solution is a task divided into exploration phase and exploitation phase. Where exploration is search for broader areas containing possible solutions, exploitation refers to concentrating on a local area to find best solution in the region. This is generally called as local solution and global solution. Whereas local solution is best in the region but may not be optimal globally. This may happen because of local optima trap. However, to continue search for better global solution by more exploration takes more execution time. Hence algorithm needs to be efficient enough to balance the exploration and exploitation. Most of the research in this field is then dedicated to developing approaches for balancing these two phases by hybridizing the algorithms, modifying the parameters, adding evolutionary operators or using one optimization technique to optimize parameters of other optimizer etc. However, it is to be noted that every algorithm will not give best solution for every problem at hand. This is in line with the 'No Free lunches' theorem of [82]. As well, though evolutionary algorithms are usable in any field, tuning of their parameters and objective functions is a crucial task.

This paper reviewed various nature inspired computing paradigm as well as various algorithms based on natural phenomenon considering the earliest works from 1943 to the latest works published in January 2020. It can be easily inferred that nature is inspiring so many algorithms for optimizing various engineering and industrial problems. Bio-inspired optimization has unlocked many techniques for solving complex optimization issues- constrained, linear, non-linear and alike. This works takes the interested reader on a detour touching many aspects of bioinspired optimization.

While writing this work it is observed that these basic methods are further studied by many researchers either for testing their applicability to other problems or for improving the performance. For example, after going through the literature based on grey wolf optimizer, it is observed that in a short span of five years GWO has been modified, hybridized and applied through more than 300 published works. This shows that there is huge scope of using these bio-inspired algorithms for solving many complex problems.

**Pravin S. Game** is a Ph.D. scholar in Computer Engineering at Shri JJT University, Jhunjhunu, Rajasthan. He received his Master of Engineering in Computer Engineering from Savitribai Phule Pune University and Bachelor of Engineering in Computer Science & Engineering from Sant Gadge Baba Amravati University. Currently, he is working at Pune Institute of Computer Technology, Pune. His research interests include data mining, big data analysis, machine learning.
Email:pravinsgame@gmail.com.

**Dr. Vinod Vaze** is Ph.D. in Computer Engineering. He is B. Tech. from I.I.T., Kanpur, and has also earned PGDFM, Diploma in Cyber Law. He is currently working in Department of Computer Science and Engineering at Shri JJT University, Jhunjhunu, Rajashtan. His research interest includes machine learning, and cyber security.

**Dr. Emmanuel M**. is Ph.D. in Computer Science and Engineering. He is M. Tech. and B. Tech. in Computer Science and Engineering. He is currently working at Pune Institute of Computer Technology, Pune. He is leading the Big Data research group at PICT. His research interest includes big data, business intelligence, medical image processing and machine learning.